# Underwater Fish Tracking for Moving Cameras based on Deformable Multiple Kernels

Meng-Che Chuang, Jenq-Neng Hwang, Jian-Hui Ye, Shih-Chia Huang, and Kresimir Williams

*Abstract*—**Fishery surveys that call for the use of single or multiple underwater cameras have been an emerging technology as a non-extractive mean to estimate the abundance of fish stocks. Tracking live fish in an open aquatic environment posts challenges that are different from general pedestrian or vehicle tracking in surveillance applications. In many rough habitats fish are monitored by cameras installed on moving platforms, where tracking is even more challenging due to inapplicability of background models. In this paper, a novel tracking algorithm based on the deformable multiple kernels (DMK) is proposed to address these challenges. Inspired by the deformable part model (DPM) technique, a set of kernels is defined to represent the holistic object and several parts that are arranged in a deformable configuration. Color histogram, texture histogram and the histogram of oriented gradients (HOG) are extracted and serve as object features. Kernel motion is efficiently estimated by the mean-shift algorithm on color and texture features to realize tracking. Furthermore, the HOG-feature deformation costs are adopted as soft constraints on kernel positions to maintain the part configuration. Experimental results on practical video set from underwater moving cameras show the reliable performance of the proposed method with much less computational cost comparing with state-of-the-art techniques.**

## I. INTRODUCTION

ESTIMATING the abundance of commercially important fish populations is critically required by studies in fisheries science and oceanography [1]. Fisheries scientists often call for the use of bottom and midwater trawls to estimate fish abundance when conducting fisheries surveys. To facilitate this with a non-extractive approach, studies has been investigated into conducting surveys by using underwater cameras that are installed with a trawl [2]. The vast amount of collected data from camera-based surveys are processed by automatic video

Manuscript received October 19, 2015; revised and accepted November 28, 2015. This work was supported by the National Marine Fisheries Services' Advanced Sampling Technology Working Group, National Oceanic and Atmospheric Administration, Seattle, WA, USA.

M.-C. Chuang and J.-N. Hwang are with the Department of Electrical Engineering, University of Washington, Seattle, WA 98195 USA (email: {mengche, hwang}@uw.edu).

J.-H. Ye and S.-C. Huang are with the Department of Electronic Engineering, National Taipei University of Technology, Taipei 10608, Taiwan (email: {t102419017, schuang}@ntut.edu.tw).

K. Williams is with the Alaska Fisheries Science Center, National Oceanic and Atmospheric Administration (NOAA), Seattle, WA 98115 USA (email: kresimir.williams@noaa.gov).

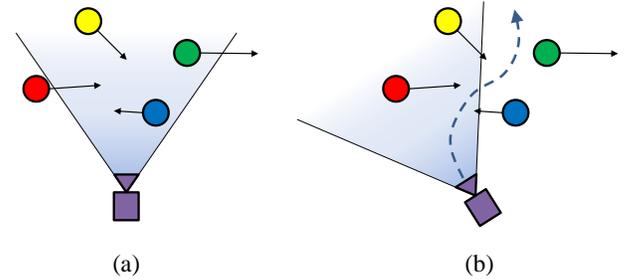

Fig. 1. Comparison of stationary and moving cameras. (a) A stationary camera covers a fixed field of view. Targets are located and tracked by maintaining a background model. (b) A moving camera has ego-motion and its covered field of view changes continuously. Background modeling techniques are no longer applicable. Camera ego-motion also introduces error when estimating target motion.

analysis technologies such as object detection and tracking [3-7]. With successful development of the required algorithms, video-based fisheries surveys not only alleviate the heavy manual labor consumption, but also enhance the spatial and temporal resolution of sampling.

In many situations, however, trawls are not applicable due to the rough surface in some habitats or the target animals tending to rest in between obstacles [8]. An alternative approach to conducting fisheries surveys is install cameras on mobile underwater vehicles such as remotely-operated vehicles (ROV) or autonomous underwater vehicles (AUV). From the aspect of video processing, the difference between moving cameras and traditional stationary cameras is introduced by camera ego-motion, as shown in Fig. 1. The main challenge posted by camera ego-motion is that existing mature background modeling techniques such as the Gaussian mixture model (GMM) [9] are no longer applicable since the field of view changes continuously [10-12]. Also, object perspective and scale can change more rapidly across time; i.e., primitive size and shape information of a tracked object is not invariant. Object tracking for moving cameras becomes even more challenging when the cameras are down in an open aquatic habitat as conducting fisheries surveys. Firstly, low image quality and ubiquitous noise in the water degrades localization accuracy. Unlike pedestrian or vehicle motion being restricted on a 2-D plane, underwater targets such as fish moves with three degrees of freedom and breaks the assumptions made by most 3-D tracking techniques. Moreover, color homogeneity and lack of background landmarks make it unable to retrieve the 3-D information from video by Visual-SLAM integrated with the structure-from-motion framework [11, 13, 14].

To address these challenges, object tracking integrated with frame-by-frame object detection has been investigated in recent years and is referred to as the *tracking-by-detection* paradigm [15-20]. For all video frames, object detection gives the patches of object's presence, which are then associated along the time to form the tracks via local or global optimization schemes. There is a large body of work on generic object detection algorithms such as the implicit shape model (ISM) [21], C⁴ [22], Regionlets [23] and deformable part model (DPM) [24]. Among these techniques, the DPM has shown its effectiveness in challenging cases and simplicity of the discriminative model comparing to later work based on deep structures [25, 26]. Based on a variant of the histogram of oriented gradients (HOG) features [27], the DPM depicts the object category by the pictorial structure framework [28], which comprises the holistic appearance and a collection of parts arranged in a deformable configuration. The detection-based paradigm has the advantage of requiring no knowledge about background appearance or target motion, and hence it is suitable for moving cameras. However, the computational cost is extremely heavy since it scans each video frame via a window with dozens of image scales, which leads to an exponential growth in the number of object proposals.

Another type of approach that can handle camera motion is kernel-based tracking [29-33]. This type of method builds a target model in terms of a color histogram where each pixel is weighted by its spatial distance to the object center. The mean-shift algorithm is then employed to efficiently find the local maximum of the feature similarity function [30]. To handle partial occlusions, a multiple-kernel approach is introduced that represents a target by more than one kernel [31]. Projected gradient is used to optimize the kernel locations under some given equality constraints. However, kernel-based tracking methods fails easily when there is a high similarity in color between the target and background or among several nearby targets.

By taking the merits from both DPM detection and multiple-kernel tracking, a novel tracking algorithm based on deformable multiple kernels to handle camera motion is proposed in this paper. In recent years there has been various part-based object tracking methods investigated in the literature [34-41]. Compared with these techniques, the method proposed in this paper has several advantages summarized as follows. 1) The proposed method integrates a given pictorial structure into kernel-based tracking and hence enables an efficient object tracking solution without training required. 2) The proposed method works successfully for cameras that moves underwater, where object tracking is much more challenging than tracking pedestrians or vehicles on the ground [11, 15-20, 34-40]. 3) By performing the kernel-based approach, the computational consumption for object tracking is significantly reduced compared to the tracking-by-detection methods [15-20]. 4) By using a hybrid feature that covers color, texture and gradient, a reliable kernel similarity evaluation yields excellent performance against the realistic underwater animals, where color homogeneity reduces the accuracy of existing tracking methods developed for pedestrian or vehicle. Recently there is an emerging body of literature for part-based object tracking methods [34-41]. Comparing with the state-of-the-art, the proposed method is advantageous in high computational efficiency, requires no training for tracking model. The proposed method is further capable of tracking genuine 3-D motion made by freely-swimming fish in contrast to most of the current tracking work on pedestrians and vehicles, which implicitly assumes the objects moves on a 2-D ground plane.

The rest of this paper is organized as follows. Section 2 introduces the background of kernel-based tracking and deformable part model. Section 3 describes the proposed deformable multiple-kernel tracking algorithm. Section 4 reports experimental results. Finally, the conclusion is given in Section 5.

## II. BACKGROUND

### A. Kernel-Based Tracking

As one of the major categories of object tracking techniques, the kernel-based method has shown its advantages in relatively low computation and robustness against non-rigid deformation [30]. The basic concept of kernel-based tracking is iteratively computing the kernel motion so that the feature distribution between the candidate and the target is best matched. This can be formulated as a search for the position that maximizes a density estimator

$$f(\mathbf{x};h) = \frac{\sum_{i=1}^{N_h} w_i k(\left\|\frac{\mathbf{x}-\mathbf{z}_i}{h}\right\|^2)}{\sum_{i=1}^{N_h} k(\left\|\frac{\mathbf{x}-\mathbf{z}_i}{h}\right\|^2)}, \tag{1}$$

where $\mathbf{x}$ is the kernel center, $\mathbf{z}_i$ is the $i$-th pixel position within the kernel, $w_i$ is the sample weight of $\mathbf{z}_i$, $k(\cdot)$ is the kernel function with a bandwidth $h$ and $N_h$ is the total number of pixels within the kernel.

In general, the target and the candidate model are represented as the probability density function of features; i.e., a color histogram where the contribution of each pixel is spatially weighted by the kernel function $k(\cdot)$. The optimal kernel position is efficiently found by applying the mean-shift algorithm [29]. More specifically, given the current kernel position $\mathbf{x}$, the new kernel position is computed by $\mathbf{x}' = \mathbf{x} + \mathbf{m}(\mathbf{x})$, where

$$\mathbf{m}(\mathbf{x}) = \frac{\sum_{i=1}^{N_h} w_i k'(\left\|\frac{\mathbf{x}-\mathbf{z}_i}{h}\right\|^2)(\mathbf{z}_i - \mathbf{x})}{\sum_{i=1}^{N_h} w_i k'(\left\|\frac{\mathbf{x}-\mathbf{z}_i}{h}\right\|^2)} \in \mathbb{R}^2 \tag{2}$$

is referred to as the *mean-shift vector*.

One major challenge to conventional kernel-based tracking is partial occlusion; i.e., a part of the target is behind some obstacles or other targets. The occluded part is not visible and thus introduces error when computing histogram similarity. In [31], the constrained multiple-kernel (CMK) tracking method is proposed. A target is represented by a number of kernels that comply with predefined spatial constraints. By associating adaptive weights to the kernels, the adverse effect introduced by partial occlusion can be compensated. The kernel motion is given by minimizing the sum of cost functions $J_i(\mathbf{x})$ subject to the constraints $C(\mathbf{x})$; i.e.,

$$\min_{\mathbf{x}} \sum_{i=1}^{n} w_i J_i(\mathbf{x}), \qquad (3)$$

$$\text{s.t. } C(\mathbf{x}) = 0, \qquad (4)$$

where $w_i$ is the weight for the $i$-th kernel. A projected gradient approach is applied to solve the above equality-constrained optimization problem.

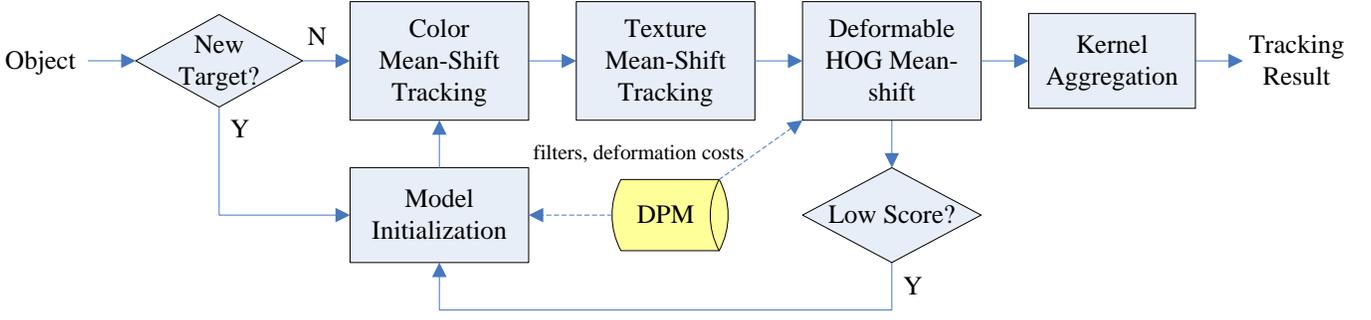

Fig. 2. Overview of the deformable multiple-kernel (DMK) tracking algorithm. Given the detected object, a multiple-kernel target model is initialized such that the kernels correspond to root and part filters from the deformable part model (DPM). The location and size of each kernel are updated via the mean-shift algorithm for spatially-weighted color histogram, texture histogram and HOG features. The deformation cost is imposed in the HOG mean-shift stage to maintain the part configuration. If the matching score in HOG mean-shift stage is low, the target model is reset. Finally, the tracking result (in terms of bounding box location and size) is produced by a kernel aggregation method.

### B. Deformable Part Model

The deformable part model (DPM) [24] has been regarded as one of the most generic and powerful object detectors even for challenging scenarios by capturing significant intra-class variations. The DPM uses discriminative training and learns object appearance based on the pictorial structure [28], which represents an object with a collection of parts in a deformable configuration.

The DPM for an object category consists of a root filter $\mathbf{F}_0$, $n$ part models $P_1, ..., P_n$ sampled at twice the resolution of the root filter, and a real-valued bias term $b$. The $i$-th part model $P_i$ is defined by a 3-tuple $(\mathbf{F}_i, \mathbf{v}_i, \mathbf{d}_i)$, where $\mathbf{F}_i$ is the part filter, $\mathbf{v}_i \in \mathbf{R}^2$ specifies the anchor position for the part relative to the root position, and $\mathbf{d}_i \in \mathbf{R}^4$ denotes coefficients of a quadratic deformation cost function for part displacement.

Detecting objects in an image is done by applying linear filtering to a feature pyramid, which consists of dense feature maps computed from each layer of a standard image pyramid. A variation of the HOG [27] is used as the feature vector. Let $H$ be a feature pyramid and $\mathbf{p} = (\mathbf{x}, l) = (x, y, l)$ denote a specific position and level in the pyramid. Define an object hypothesis as the location of each filter in the feature pyramid,

$\mathbf{z} = (\mathbf{p}_0, ..., \mathbf{p}_n)$, with the constraint that all parts are placed at twice the resolution of the root. The detection score of a hypothesis $s(\mathbf{p}_0, ..., \mathbf{p}_n)$ is then given by the sum of all filter responses at their respective position subtracted by the deformation cost of each part, and plus the bias term; i.e.,

$$s(\mathbf{p}_0, ..., \mathbf{p}_n) = \sum_{i=0}^{n} \mathbf{F}_i \cdot \phi(H, \mathbf{p}_i) - \sum_{i=1}^{n} \mathbf{d}_i \cdot \phi_d(d\mathbf{x}_i) + b, \qquad (5)$$

where $\phi(H, \mathbf{p}_i)$ denotes the feature vector obtained from the location $\mathbf{p}_i$ in the feature pyramid, $d\mathbf{x}_i = (dx_i, dy_i) = \mathbf{x}_i - (2\mathbf{x}_0 + \mathbf{v}_i)$ denotes the displacement of the $i$-th part with respect to its anchor position (note the factor 2 is due to double-resolution sampling of the part models), and $\phi_d(d\mathbf{x}) = (dx^2, dx, dy^2, dy)$ is the deformation features that express the displacement in the quadratic form. The bias $b$ is introduced to accommodate the constant term when training the final binary classifier for $s(\mathbf{p}_0, ..., \mathbf{p}_n)$. A latent SVM formulation is proposed to train the DPM parameters $\{\mathbf{F}_0, P_1, ..., P_n, b\}$ for the object category of interest.

### III. DEFORMABLE MULTIPLE-KERNEL TRACKING

Inspired by multiple-kernel representation of an object [31], the proposed deformable multiple-kernel (DMK) tracking algorithm regards each part model as a kernel. Like the predefined constraints that bind the kernels in the CMK formulation, the DMK algorithm adopts the deformation costs to restrict the displacement of kernels during tracking. For each video frame, the proposed algorithm iteratively shifts the kernels based on weighted color histogram, texture histogram and HOG. As a result, the DMK approach takes advantage of not only low computational cost from the kernel-based tracking but also the robustness of object localization from the DPM detection. An overview of the proposed deformable multiple-kernel tracking algorithm is shown in Fig. 2.

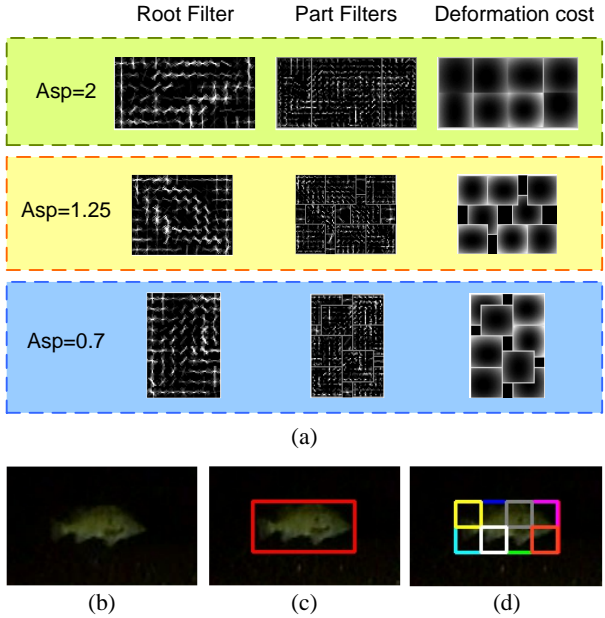

|  | Root Filter | Part Filters | Deformation cost |
|---|---|---|---|
| Asp=2 | | | |
| Asp=1.25 | | | |
| Asp=0.7 | | | |

(a)

(b)   (c)   (d)

Fig. 3. Deformable multiple-kernel model (a) the fish DPM with 3 components, each of which is presented with a specific aspect ratio of fish body to capture different viewing perspective. One DPM component consists of a root filter and 8 part filters (sampled at twice the resolution of root filter); (b) a target object in the video; (c) root kernel corresponding to the root filter; (d) part kernels corresponding to the part filters and located according to the deformation cost map. This figure is best viewed in color.

### A. Models

In practice, the DPM represents an object category by a mixture of star models, which consist of a number of components. Each component is trained by a group of positive samples with similar bounding box aspect ratio, which serves as a simple indicator of intra-class variation in perspective. An example of the mixture DPM for fish is shown in Fig. 3(a). As a result, the DMK model is initialized by using the most suitable component as follows.

The DMK model is composed of $n+1$ kernels. First, the root kernel is set identically to the object bounding box. Let $R_i$ and $C_i$ be the number of cell rows and columns for the $i$-th filter. We then select the DPM component with the root filter aspect ratio $C_0 / R_0$ most similar to the bounding box aspect ratio $w / h$. Based on the selection, each part kernels is then placed at its anchor position. The size of each part kernel is scaled by the size ratio between the part and root filter such that $h_i / h_0 = R_i / R_0$ and $w_i / w_0 = C_i / C_0$ for each part kernel size $(w_i, h_i)$. An example of multiple kernels for an object is shown in Fig. 3(b)-3(d). For kernel features, we use the color histogram, texture histogram and HOG, which are described in the following subsections.

### B. Tracking by Color and Texture

In kernel-based tracking [30], a target is represented by a color histogram where each pixel's contribution is spatially weighted by a kernel function. The kernel motion is efficiently computed by maximizing the histogram similarity between a candidate model and the given target model via the mean-shift algorithm.

The weakness of this approach is that the similarity metric is not reliable when the target color is not distinct from the environment (background and other targets). To improve this, the texture attribute is also utilized as an additional feature. We use the rotation-invariant uniform type of local binary patterns (LBP) histogram proposed in [42]. This type of pattern is derived such that the feature descriptor is not only invariant to rotation but also robust to noisy high-frequency patterns. Similar to color, an LBP histogram is constructed with the spatial weight given by the kernel function. Each kernel is then moved to its optimal location by applying the mean-shift algorithm separately. The mean-shift algorithm is iterated until the maximum iteration ($T_{\text{color}}$ for color and $T_{\text{texture}}$ for texture) is reached. The texture feature is applied after the original color histogram feature so as to form a coarse-to-fine framework: object location is first obtained from larger range with a certain level of accuracy based on the color attribute and then refined locally with a much higher precision by imposing the informative texture features.

### C. Deformable HOG Mean-Shift

After the color and texture kernel tracking, part kernels are likely to move away from their anchor positions. To restore the object part configuration, these part kernels are further shifted to optimize the HOG similarity while being bound with the root kernel by the deformation cost functions, which are similar to the imposed constraints used in the CMK tracking. The effect of restoring part configuration by the proposed deformable HOG mean-shift algorithm is demonstrated in Fig. 4.

To ensure the accuracy of HOG feature matching, we first calculate the object scale; i.e., the feature pyramid level where the object is detected. Following the notations in Section III.A, given a root filter with $R_0$ rows and $C_0$ columns of cells, the image pyramid level of the object's presence is

$$l_0 = \left\lfloor \lambda \log_2(\min\{w_0 / kC_0, h_0 / kR_0\}) \right\rfloor, \quad (6)$$

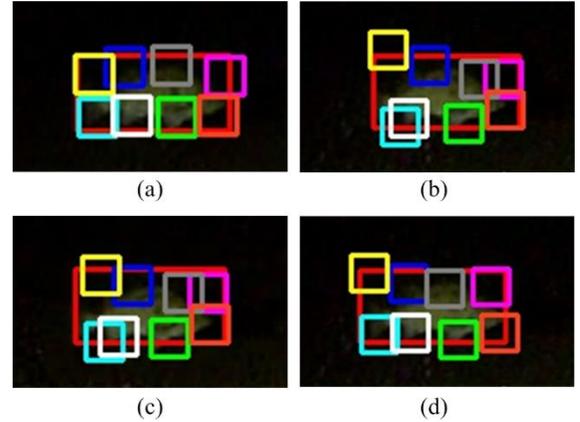

(a)   (b)

(c)   (d)

Fig. 4. The root kernel (red) and part kernels (other colors): (a) from the previous frame; (b) after color mean-shift; (c) after texture mean-shift; (d) after HOG mean-shift, which considers the deformation costs. Note the restoration of part configuration according to the DPM (such as the cyan and white kernels on the bottom left corners). This figure is best viewed in color.

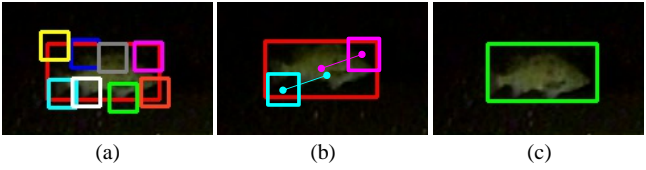

|  (a)  |  (b)  |  (c)  |

Fig. 5. Example of kernel aggregation: (a) the deformable multiple-kernel model after tracking; (b) the root kernel (red) and two part kernels (cyan and magenta), whose anchor vectors and projected object centers given by (9) are illustrated; (c) final bounding box. This figure is best viewed in color.

where $(w_0, h_0)$ is the detected width and height of the object and $k$ denotes the HOG square cell size in pixels. Following the standard setting in [27] we set $k = 8$ pixels. In practice, there are $\lambda = 10$ levels per octave in the image pyramid during object detection. For part kernels, features are extracted at twice the resolution of the root kernel level; i.e., $l_i = l_0 - \lambda$ for $i = 1, \ldots, n$.

Below we following the notations in Section II.B. For the $i$-th kernel ($i = 0$ for root and $i = 1, \ldots, n$ for parts), let $\mathbf{p}_i = (\mathbf{x}_i, l_i) = (x_i, y_i, l_i)$ specify a position and pyramid level, $\phi(H, \mathbf{p}_i)$ denote the corresponding HOG features, $d\mathbf{x}_i = (dx_i, dy_i)$ denote the kernel displacement from its anchor position, and $\phi_d(d\mathbf{x}_i) = (dx_i^2, dx_i, dy_i^2, dy_i)$. Given the DPM filter vector $\mathbf{F}_i$ and deformation coefficients $\mathbf{d}_i$, the HOG similarity for the $i$-th kernel is given by

$$s(\mathbf{p}_i) = \mathbf{F}_i^* \cdot \phi^*(H, \mathbf{p}_i) - (1 - \delta_i)(\mathbf{d}_i^* \cdot \phi_d^*(d\mathbf{x}_i)), \quad (7)$$

where $\delta_i$ denotes the Kronecker delta function at $i = 0$, and $(\cdot)^*$ denote $L^2$-normalized vectors. Vector normalization is imposed to eliminate the magnitude difference between two terms. The mean-shift algorithm is applied to update the location of each kernel based on the HOG similarity $s(\mathbf{p}_i)$. The iteration continues until the maximum iteration $T_{HOG}$ is reached.

### D. Kernel Aggregation

The object bounding box is determined by the root and part kernels after the kernel-based tracking and deformable HOG mean-shift algorithms. We integrate the root kernel and the part kernels to obtain the final position of the bounding box as follows. Let $\mathbf{c} \in \mathbf{R}^2$ denote the original bounding box center. We define the projected object center from the $i$-th part kernel by its current position $\mathbf{x}_i$ and anchor vector $\mathbf{v}_i$; i.e., $\mathbf{c}_i = (\mathbf{x}_i - \mathbf{v}_i)/2$. The new center position is then given by

$$\mathbf{c}' = \alpha \mathbf{c}_{root} + (1 - \alpha)\mathbf{c}_{part}, \quad (8)$$

where $\alpha \in [0,1]$, $\mathbf{c}_{root} = \mathbf{x}_0$ is the root kernel center, and

$$\mathbf{c}_{part} = \mathbf{c} + \frac{\sum_{i=1}^{n} s(\mathbf{p}_i)(\mathbf{c}_i - \mathbf{c})}{\sum_{i=1}^{n} |s(\mathbf{p}_i)|} \quad (9)$$

is a weighted average projected bounding box center. The weight $s(\mathbf{p}_i)$ is directly adopted from (7) as an indication of how well the candidate match the target. A higher weight value corresponds to higher confidence of this part kernel. The absolute value in the denominator ensures the capability of handling negative similarity values. In this way, the final object position is robust against partial occlusions or low discrimination of the feature around the neighborhood. An example of computing the bounding box of the target by the kernel aggregation approach is illustrated in Fig. 5.

### E. Scale Adjustment

Targets captured by a moving camera are likely to undergo rapid change in scale throughout the tracking lifespan. The proposed algorithm updates the scale of tracks by adopting the mechanism based on the derivative of density estimate $f$ with respect to the kernel bandwidth $h$ [31]. Here the target scale is represented by the size of target bounding box. Let $s_k \in \mathbf{R}^2$ be the target scale at frame $k$, and $\beta$ denote the update step size. The target scale at frame $k+1$ is then updated by

$$s_{k+1} = (1 + \Delta s)s_k = \left(1 + \beta \frac{\nabla f(h)}{f(h)}\right)s_k, \quad (10)$$

## IV. EXPERIMENTAL RESULTS

### A. Implementation Details

We adopt the public source code of DPM object detection provided by the original authors [43]. It is worthwhile to note here that the DPM is a well-known detection algorithm, which merely finds object instances in a single still image. In contrast, the proposed method is a tracking algorithm, which not only locates multiple objects in a single image but also labels the objects along time (across consecutive images or video frames) even some objects entering or exiting halfway. Furthermore, as shown in Fig. 2, the DPM only provides a representation (filters and deformation costs) of the object type of interest to the proposed DMK tracking method, which then use the representation to perform the actual tracking pipeline.

All values of parameters in the experiments are determined empirically if not specified. The number of parts in the DPM is determined automatically by the algorithm during the training stage. The mixture DPM is employed to enable capturing targets viewed in different perspectives throughout the videos. The number of perspective components in the mixture DPM is set as $M = 3$, which is chosen empirically based on the trade-off between robustness against intra-class variations and simplicity of trained models.

The color histogram is built in the HSV color space to reduce the influence of brightness discrepancy between the target and candidates. The hue channel is divided equally into 15 bins. The saturation and value channels are divided equally into 8 bins respectively. For texture features, we follow the settings from [42] and generate a 10-bin histogram of the RIU-LBP for each kernel. The HOG features in each cell is represented by a 9-bin histogram. For mean-shift procedures, the maximum iterations are set as $T_{color} = 5$, $T_{texture} = 5$ and $T_{HOG} = 3$. K-L distance is used as the similarity metric for color and texture histograms. In kernel aggregation, we set the root center weight $\alpha = 0.5$ empirically. The proposed algorithm is not very dependent on the value of scale update step size $\beta$. A high value such as 10000 should be enough for most cases. We set the criteria for reinitializing the DMK model as either the total HOG matching score of all kernels is lower than 0.2 or the scores are decreased for 3 consecutive frames. Finally, a linear Kalman filter is applied in every video frame to smooth the target trajectory.

TABLE I
AVERAGE TRACKING ERROR (PIXELS)

| Dataset | MS [21] | CMK [22] | DMK-clr-hog | DMK-clr-tex-hog |
|---------|---------|----------|-------------|-----------------|
| ROVVTS_1 | 101.45 | 48.51 | 27.31 | **8.61** |
| ROVVTS_2 | 57.80 | 39.16 | 22.45 | **8.11** |
| AFSCUW_1 | 89.12 | 18.24 | 9.84 | **7.29** |
| AFSCUW_2 | 27.52 | 56.18 | 6.91 | **5.56** |
| HAUL1010_1 | 59.91 | 28.50 | 37.00 | **7.38** |
| HAUL1010_2 | 200.37 | 76.67 | 24.14 | **11.67** |

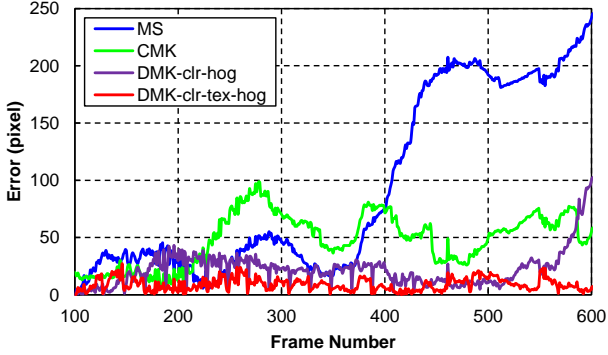

(a)

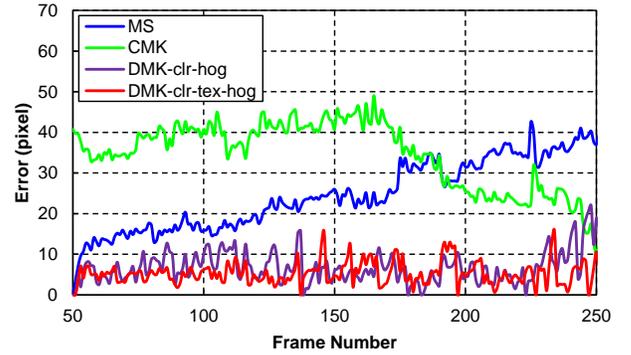

(b)

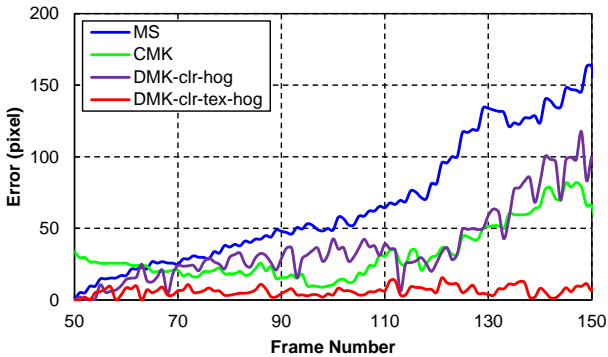

(c)

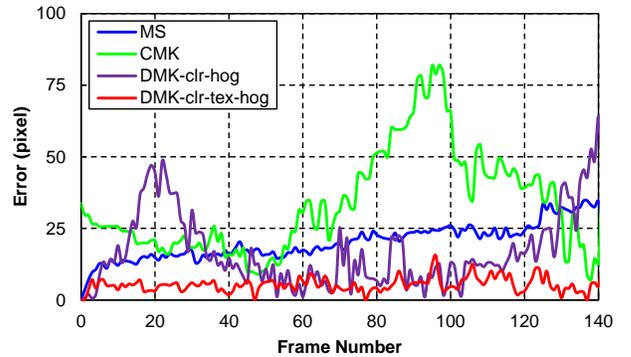

(d)

Fig. 6. Errors of target center in pixels vs. frame number of tested datasets. (a) ROVVTS1_1; (b) AFSCUW_2; (c) HAUL1010_1; (d) HAUL1010_2.

### B. Tracking Accuracy

Tracking objects for a moving camera is a rather new category that differs from traditional tracking for a stationary camera, or objects in a moving camera but with motion restricted to a 2-D plane (ground plane) such as pedestrians and vehicles. To the authors' best knowledge, there are no existing public datasets that serves as benchmark for this category of tracking. As a result, we tested the proposed method by its tracking accuracy on several large-scale underwater mobile video datasets collected by the Alaska Fisheries Science Center, National Oceanic and Atmospheric Administration (NOAA). The NOAA datasets used in this paper include highly-diverse underwater mobile video sequences collected from a wide range locations and habitats with a variety of target types, and thus provides a representative testbed for evaluation. Each dataset consists of over 40 minutes of video recorded at a frame rate of 29 frames per second (fps) in marine habitats in the Alaska Gulf and Pacific Ocean. The cameras were either mounted on a remotely operated underwater vehicle (ROV) or passively towed by a vessel. Since the proposed method uses an object representation based on the deformable part model (DPM), which has shown its outstanding performance in detection, the tracking performance is supposed to be the same on all categories of targets given the corresponding DPM is properly trained and used as described in this paper.

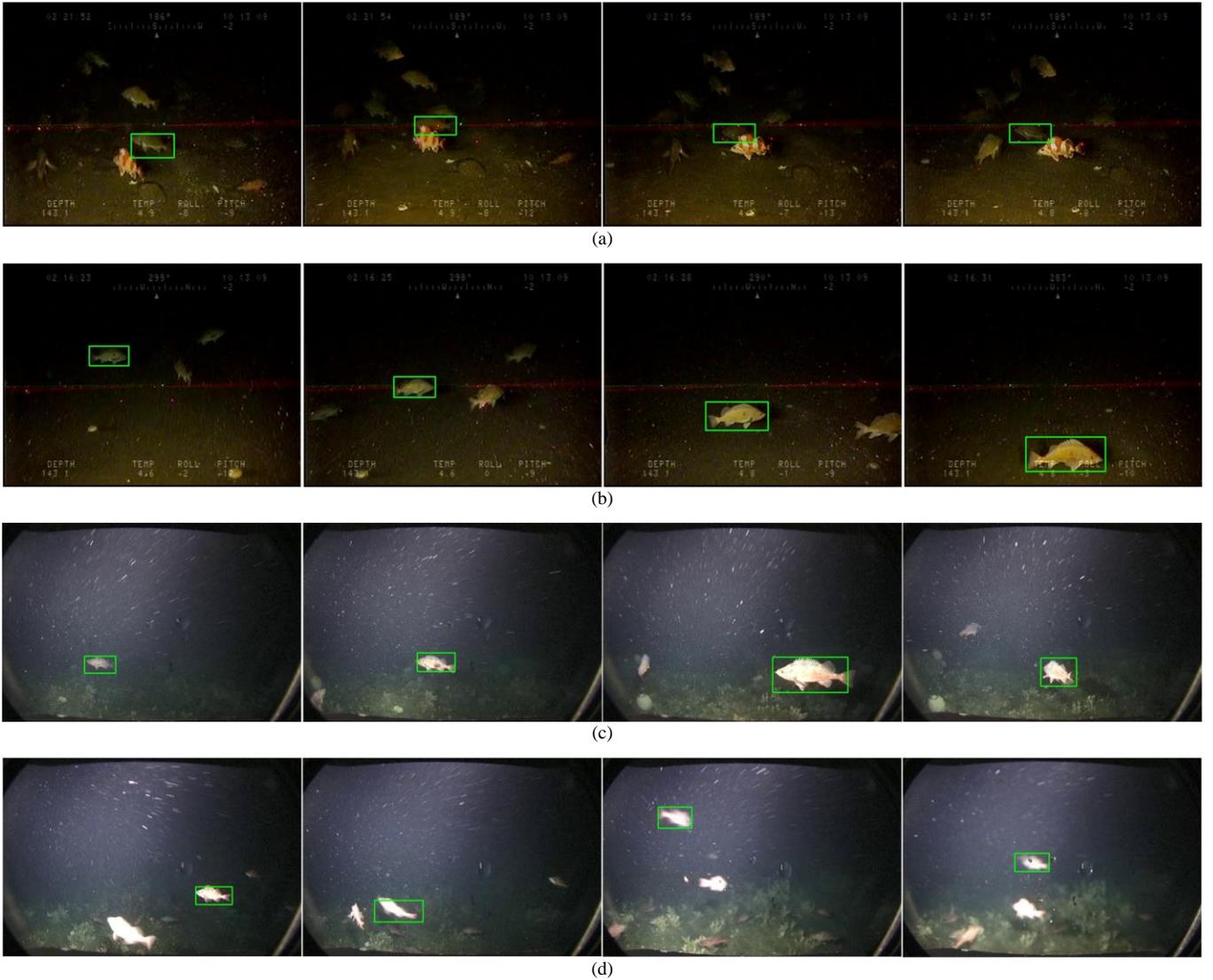

Fig. 7. Results of a specific target using the proposed DMK tracking algorithm in the tested datasets. (a) ROVVTS1_1; (b) AFSCUW_2; (c) HAUL1010_1; (d) HAUL1010_2. Partial occlusion of the target occurs in (a), while abrupt change in speed and sharp turns of target motion can be observed in (c) and (d).

The targets to be tracked are live fish, which are considered challenging due to the frequent variation in perspective and high body deformation during swimming. For each test the target of interest is located manually by the bounding box in the first video frame. The performance is evaluated by the tracking error; i.e., the spatial distance measured in pixels between the center of target bounding box and the manually-labeled ground truth. Some clips of the tested video associated with the simulations reported in this paper can be watched online.[1]

Since most assumptions on object location and motion do not hold in the underwater scenario, it is unfair to compare the proposed method with most part-based algorithms that are designed for tracking pseudo-3D targets such as pedestrians or vehicles. On the other hand, the proposed method belongs to the kernel-based type in object tracking techniques. It is therefore compared with the traditional mean-shift tracking (MS) [30] and constrained multiple-kernel tracking (CMK) method [31]. In CMK realization we represent a target by two kernels of identical size and are aligned horizontally. The spatial distance between two kernels is regarded as the constraint. To demonstrate the effectiveness of adopting texture features, we also report the results of the proposed method without taking into account the LBP histograms.

Table I reports the average tracking error of these techniques throughout the video frames on each of the tested datasets. The proposed DMK tracking method significantly outperforms others in all of the video datasets. By taking texture attributes into account, the proposed method successfully tracks the targets with little to no discriminative color distributions comparing to its vicinity. The deformation cost functions, which effectively retains the configuration of parts, also help determine the object position with better accuracy. On the contrary, the MS method uses only one kernel for the target, and hence suffers from drifting when occlusion occurs. The CMK method handles occlusion better than the MS method does, but still gives the results with large error due to the inflexible spatial restrictions and dependency on only color

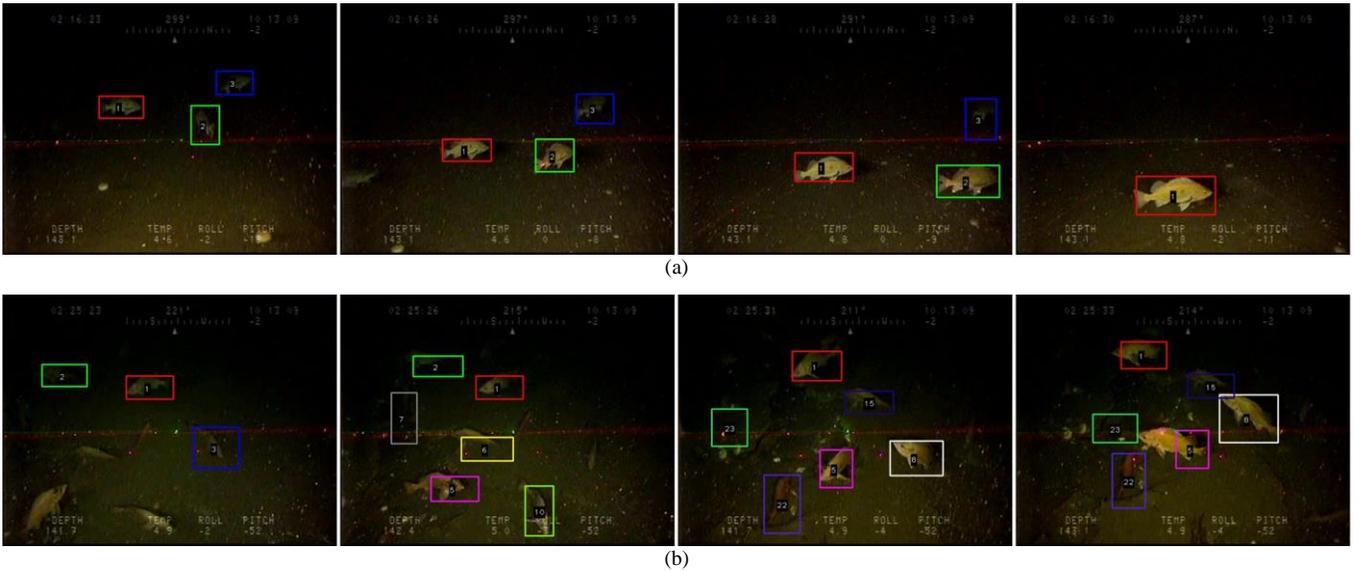

(a)

(b)

Fig. 8. Results of a tracking multiple targets simultaneously: (a) ASFCUW_2; (b) ROVVTS_3. This figure is best viewed in color.

histogram features.

Time plots of tracking error (error versus frame number) for all techniques are also shown in Fig. 6. From these time plots one can observe the robustness of the propose method in extreme scenarios. For instance, in Fig. 6(a) the target is partially occluded from frame #340 to #440, which can be seen in Fig. 7(a). The proposed DMK method successfully recovers the target's trajectory because the kernel aggregation scheme suppresses those part kernels with lower tracking confidence. Another major challenge to tracking live fish is the abrupt changes in velocity. As shown in Fig. 7(c) and (d), the proposed method demonstrates its capability of tracking targets with sudden acceleration and sharp turns by adopting texture and HOG features, which are more reliable in the noisy underwater environment.

### C. Multiple-Target Tracking

The proposed method is further tested for multiple-target tracking; i.e., all targets visible in the video are tracked simultaneously. Here the underwater moving camera datasets collected by NOAA are again used as the testbed, and the DPM object detection is performed to locate the targets since background modeling techniques are not applicable to moving cameras. For each object, an independent DMK model is initialized and updated via the proposed method as what is applied in single-target tracking. The results of multiple-target tracking using the proposed DMK method are shown in Fig. 8. The proposed method successfully tracks all objects that are visible in the video despite the color ambiguity and occlusion against some of the objects. In addition to the MS and CMK methods from previous experiment, the proposed method is also compared with Berclaz et al. [16], which is the state-of-the-art of tracking-by-detection methods. The average tracking error for all targets are reported in Fig. 9. The proposed method still outperforms the MS and CMK methods. However, the performance is slightly worse than the detection-based method in 4 out of 6 datasets. This is not surprising since the detection-based method uses the object detection results directly and thus gives the most accurate results as long as object detection is reliable. The exception can be found in datasets ROVVTS_1 and AFSCUW_1, where frequent partial occlusion leads to degraded accuracy in object detection. Some representative results of multiple-target tracking by the proposed method are shown in Fig. 8.

The advantage of the proposed method over the tracking-by-detection methods, however, is the computational requirement. This is credited to high efficiency of kernel-based tracking techniques. To verify this, we compare the CPU time consumption for updating the location of the target for one video frame. Table II reports the average computation time for tracking one target using several techniques, including the tracking-by-detection approaches as well as the kernel-based approaches. From Table II, the approach by Berclaz et al. takes the highest computation due to the multi-scale scanning by the object detector. The MS method uses the efficient kernel-based tracking algorithm and takes the least computation time. However, it leads to worse tracking accuracy in complex moving-camera cases as demonstrated in Section IV.B. The proposed DMK tracking technique, which combines the advantages from object detector and kernel-based tracking, achieves high performance in challenging moving camera scenarios while requires low computational cost by using the efficient mean-shift algorithm.

### D. Discussion

One major factor to the tracking accuracy in the proposed DMK tracking algorithm is the weight during kernel aggregation introduced in (8). The effect of the kernel aggregation weight is balancing the influence on the target position by the root kernel and part kernels after a series of deformable part shifting. When the target is well visible in the field of view, part kernels are mostly well aligned with the expected position given by the anchors. In this case, either the root kernel position or the position projected by part kernels serves as a good estimate of the object position. If partial occlusion occurs, the root kernel position is likely off the real target position since there exists significant error when calculating the histogram similarity. Moreover, some of the part kernels are unable to reach the positions with high matching score since the corresponding parts are invisible or the positions are constrained by the deformation cost functions. As a result, performing kernel aggregation with a higher weight value leads to a target position contributed less by the part kernels and more robust against deformation. On the other hand, a lower weight gives more accurate estimate of target position when the object is highly deformed, but very sensitive if any part is occluded.

TABLE II
COMPUTATIONAL CONSUMPTION

| Methods | MS [21] | CMK [22] | Berclaz [12] | proposed |
|---------|---------|----------|--------------|----------|
| Avg. time per frame (sec) | 0.235 | 0.263 | 7.834 | 1.069 |

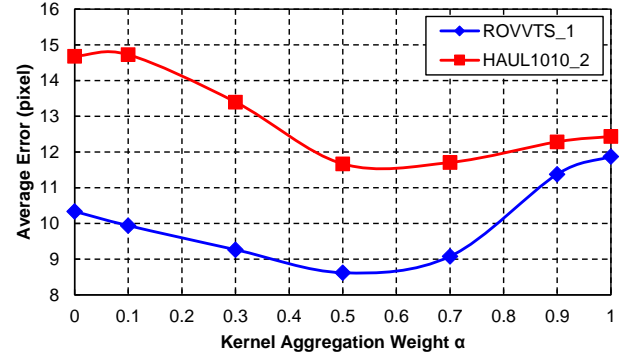

Fig. 10. Average tracking error by the proposed method vs. kernel aggregation weight in the tested dataset.

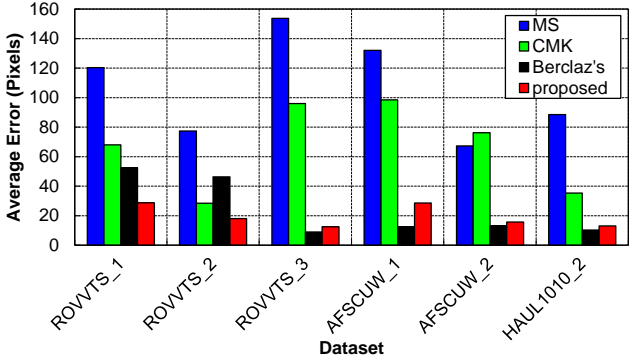

Fig. 9. Average tracking error of tested datasets for multiple-target tracking.

To visualize the effect of the kernel aggregation weight, we tested the proposed algorithm with different weight values with the underwater video datasets. Single target tracking is performed with the target manually localized in the first video frame. Other simulation settings are the same as what is described in Section IV.B. The tracking performance with $\alpha = 0$ and $\alpha = 1$ are also reported as references to demonstrate the extreme cases; i.e., estimating the target location by only the part kernels or only the root kernel.

The results are illustrated in Fig. 10. One can see that the tracking performance in two of the datasets are actually not very sensitive to the aggregation weight between the root and part kernels. The range of error in two datasets are within 4 pixels, which is only 5% of the diagonal length for a typical target bounding box. This implies that the root kernel center and the "expected" center by part kernels are very close to each other as well as the ground truth. In dataset ROVVTS_1, a serious partial occlusion of the target exists for approximately 100 frames. By the proposed kernel aggregation scheme, the HOG matching score is taken into account for each part kernel. The matching scores from occluded parts are much lower than the visible ones. As a result, the estimate of target center is robust against partial occlusion by suppressing the parts with lower confidence in matching. Dataset HAUL1010_2 presents another challenging scenario, namely tracking a target with abrupt motion. High deformation is often observed when the target is making sharp turns or change of speed. In this case the part kernels may not give a consistent estimate of the target center. This explains the lower performance achieved by using a small value for $\alpha$, where the part kernel locations dominate the final target center in (8).

## V. CONCLUSION

In this paper, a novel object tracking technique for moving cameras based on deformable multiple kernels is proposed for the challenging scenario of tracking a school of live fish from an underwater camera with ego-motion. A set of kernels is employed to represent not only the holistic object but also its local parts in terms of color histogram, texture histogram and HOG features. Integrating the deformable part model (DPM) widely used for object detection to the mean-shift optimization algorithm, the proposed method successfully combines the advantages from both techniques: the DPM introduces gradient features and part deformation costs to facilitate multiple-kernel tracking, while the mean-shift algorithm significantly reduces the computations required by typical tracking-by-detection paradigms. Experimental results show the proposed method outperforms both the traditional kernel-based tracking approaches and recent tracking-by-detection methods for tracking one or multiple live fish in challenging underwater videos, and thus provides an efficient solution to video-based fisheries surveys conducted with dynamic cameras.